\documentclass[11pt]{article}
\usepackage{paclic34}
\usepackage{algorithm}
\usepackage{algpseudocode}
\usepackage{amsmath}
\usepackage[utf8]{inputenc}
\usepackage[vietnamese=nohyphenation]{hyphsubst}
\usepackage[vietnamese,english]{babel}
\usepackage{times}
\usepackage{latexsym}
\usepackage{amsmath}
\usepackage{multirow}
\usepackage{url}
\usepackage{float}

\usepackage{adjustbox}
\usepackage{placeins}

\title{A Simple and Efficient Ensemble Classifier Combining Multiple Neural Network Models on Social Media Datasets in Vietnamese}

\author{
  Huy Duc Huynh \\ 
  University of Information Technology \\ 
  VNU-HCM, Vietnam \\
  {16520508@gm.uit.edu.vn} \\\And
  Hang Thi-Thuy Do \\ 
  University of Information Technology \\ 
  VNU-HCM, Vietnam \\
  {16520339@gm.uit.edu.vn} \\\AND
  Kiet Van Nguyen \\ 
  University of Information Technology \\ 
  VNU-HCM, Vietnam \\
  {kietnv@uit.edu.vn} \\\And
  Ngan Thuy-Luu Nguyen \\ 
  University of Information Technology \\ 
  VNU-HCM, Vietnam \\
  {ngannlt@uit.edu.vn} \\
  \newline
 }

\date{}

\begin{document}
\maketitle
\begin{abstract}
Text classification is a popular topic of natural language processing, which has currently attracted numerous research efforts worldwide. The significant increase of data in social media requires the vast attention of researchers to analyze such data. There are various studies in this field in many languages but limited to the Vietnamese language. Therefore, this study aims to classify Vietnamese texts on social media from three different Vietnamese benchmark datasets. Advanced deep learning models are used and optimized in this study, including CNN, LSTM, and their variants. We also implement the BERT, which has never been applied to the datasets. Our experiments find a suitable model for classification tasks on each specific dataset. To take advantage of single models, we propose an ensemble model, combining the highest-performance models. Our single models reach positive results on each dataset. Moreover, our ensemble model achieves the best performance on all three datasets. We reach 86.96\% of F1-score for the HSD-VLSP dataset, 65.79\% of F1-score for the UIT-VSMEC dataset, 92.79\% and 89.70\% for sentiments and topics on the UIT-VSFC dataset, respectively. Therefore, our models achieve better performances as compared to previous studies on these datasets.
\end{abstract}

\section{Introduction}

The rapid development of science and technology in the world has created a vast amount of data. In particular, the growth of social networks continuously creates a huge amount of comments and posts which are valuable sources to exploit and analyze in the digital era. Text classification is a prerequisite for such works as analyzing user opinion in the network environment, filtering and removing malicious information, and detecting criminal risk. With great potential, text classification has attracted much attention from experts in the natural language processing community worldwide. In English, we easily search for a range of text classification publications in many fields. However, relatively few researches have been done on Vietnamese text. Most published articles focus on binary classification. However, a large amount of information today requires analysis in many more aspects (multi-label or multi-class). The lack of knowledge and techniques for the Vietnamese language makes us decide to conduct this research to classify multi-class text for Vietnamese social media datasets. These datasets are provided from the VLSP share-task and publications on text classification. In particular, there are various social media textual datasets such as UIT-VSMEC for emotion recognition ~\cite{ho2019emotion} and UIT-VSFC for students' feedback classification ~\cite{van2018uit} and HSD-VLSP for hate speech detection ~\cite{vu2019hsd}. These are the datasets with multi-label and imbalance between the labels that have been published recently. They are suitable for the requirements that we would like to study.

The emergence of deep neural networks ~\cite{liu2017deep} and word embeddings have made text classification more efficient. Pre-trained word embeddings accurately capture semantics to assist deep learning models improve the efficiency of classification. In this study, we implement deep learning models such as CNN ~\cite{kim2014convolutional}, LSTM ~\cite{hochreiter1997long} and their variants to solve classification problems. Besides, we implement the BERT model ~\cite{devlin2018bert}, which is a state-of-the-art model in many natural language processing tasks in recent years. BERT is trained through the transformer’s two-dimensional context (a neural network architecture based on the self-attention mechanism to understand languages). BERT is in contrast to previous deep learning models that looked at a text sequence from left to right or combined left-to-right and right-to-left training. To improve the word representation, we create a normalized words dictionary, which helps recognize words included in pre-trained embedding but is not represented due to misspellings.

As a result, CNN model combined with fastText's pre-trained embedding \cite{grave2018learning}, has been remarkably performance on Vietnamese social media datasets. Our study also proves the efficiency of BERT on Vietnamese students' feedback dataset. Besides, we combine single models to increase the efficiency of the classification. As a result, our ensemble model accomplishes higher results than the single model. Compared to previous studies done on the datasets, our models achieve better results.

\section{Related Work}
Nowadays, many organizations realize the importance of sentiment analysis for consumer's feedback. Through this feedback, they can evaluate the quality of their services or products and devise appropriate strategies. In order to predict the genre and rating of films through viewer ratings, Varshit battu and his collaborators \cite{Varshit} conducted research on viewer comment data collected from many websites. They implemented various classification methods on their dataset to evaluate the effectiveness of methods. As a result, the CNN model achieved high results in many different languages.

The detection of emotions in texts has become an essential task in natural language processing. Su et al. \shortcite{Su} studied the text emotional recognition problem based on semantic word vector and emotional word vector of the input text. Their proposed method used the LSTM model for emotion recognition by modeling the input text's contextual emotion evolution. They use five-fold cross-validation to evaluate the performance of their proposed method. As a result, their model achieved recognition accuracy of 70.66\% better than the CNN-based method which was implemented in the same dataset. 

In addition, hate speech detection is increasingly concerned because of the explosion of social networks. There has been an amount of successful research in this field. To complete the offensive task of categorizing tweets that were announced by SemEval competition in 2019. Nikolov and Radivchev. \shortcite{nikolov-radivchev-2019-nikolov} used different approaches and models towards offensive tweet classification. Their paper presented pre-processing data methods for tweets as well as techniques for tackling imbalanced class distribution in the provided test data. Their experiments show that the BERT model proved its outstanding advantages in text classification. Not only did it outperform conventional models on the validation set, but also based on the results from the test set, it did not cause the over-fitting issue. 

In Vietnam, there have been some studies efforts for text classification tasks, as well as contributing Vietnamese data for the research community. Pham et al. \shortcite{pham2017nnvlp} announced a neural network-based toolkit namely NNVLP for essential Vietnamese language processing tasks, including part-of-speech tagging, chunking, named entity recognition. This toolkit achieved state-of-the-art results on these three tasks. With the two of UIT-VSMEC \cite{ho2019emotion} and UIT-VSFC \cite{8606837} datasets we used in this study, their authors performed classification tasks using a variety of deep learning methods. On the  UIT-VSMEC dataset, Ho et al. \shortcite{ho2019emotion} used Random Forest, SVM, LSTM, and CNN models to classify emotions of comments. They achieved 59.74\% with seven labels and 66.48\% with six labels by using the CNN model. With the  UIT-VSFC dataset, Nguyen et al. \shortcite{8606837} gained the highest result by the BiLSTM model with 92.03\% on sentiment and 89.62\% on the topic label.  

The above studies have shown the superiority of the deep learning models in text classification, which is the premise for us to apply them to the Vietnamese datasets in this study. By modifying the models and implementing new models, we aim to bring better results for these Vietnamese social media datasets. 

\section{Datasets}
In this task, we conducted experiments on classification methods on three Vietnamese datasets, including UIT-VSMEC \cite{ho2019emotion}, UIT-VSFC \cite{van2018uit}, and HSD-VLSP \cite{vu2019hsd}. UIT-VSMEC \cite{ho2019emotion} is provided by Ho et al, the items in this dataset are the Vietnamese’s comments from social networks. This dataset contains exactly 6,927 emotion annotated sentences with seven emotion labels: ENJOYMENT, SADNESS, ANGER, SURPRISE, FEAR, DISGUST, and OTHER. UIT-VSFC is provided by Nguyen et al. \shortcite{van2018uit}. This dataset was constructed from students' responses from a university for the sentiment classification task. This dataset includes 16,175 items with two classification parts that are important: sentiments and topics. Each sample of the training dataset is assigned one of three sentiment labels POSITIVE, NEGATIVE, or NEURAL and assigned one of four topics’ labels LECTURERS, TRAINING PROGRAM, FACILITIES, OTHER. The HSD-VLSP dataset is a Vietnamese comments dataset about Hate Speech Detection on social networks provided by the VSLP 2019 shared-task \cite{vu2019hsd}. This dataset contains the comments and posts on Facebook social networks, including 25,431 items. Each data line of the training dataset is assigned one of three labels CLEAN, OFFENSIVE, or HATE. Table 1 shows examples for each dataset and its characteristics.

\begin{otherlanguage*}{vietnamese}
\centering
\begin{table*}[!htbp]
\begin{adjustbox}{width=\columnwidth*2,center}
\begin{tabular}{cllrll}
\hline
\multirow{2}{*}{\textbf{Datasets}}                       & \multicolumn{2}{c}{\multirow{2}{*}{\textbf{Labels}}} & \multicolumn{1}{c}{\multirow{2}{*}{\textbf{\begin{tabular}[c]{@{}c@{}}Percentage \\ (\%)\end{tabular}}}} & \multicolumn{2}{c}{\textbf{Examples}}                                                               \\ \cline{5-6} 
                                                         & \multicolumn{2}{c}{}                                 & \multicolumn{1}{c}{}                                                                                     & \multicolumn{1}{c}{\textbf{Vietnamese sentences}} & \multicolumn{1}{c}{\textbf{English sentences}} \\ \hline
\multirow{7}{*}{\textbf{UIT-VSMEC}}                      & \multicolumn{2}{l}{ENJOYMENT}                        & 28.36                                                                                                     & Tốt quá!                                          & Very good!                                      \\ 
                                                         & \multicolumn{2}{l}{SADNESS}                          & 19.31                                                                                                     & Nay buồn!                                         & Today, I feel sad!                              \\ 
                                                         & \multicolumn{2}{l}{ANGER}                            & 16.59                                                                                                     & Im đi!                                             & Shut up!                                        \\ 
                                                         & \multicolumn{2}{l}{SURPRISE}                         & 6.92                                                                                                      & Đẹp trai mà bị đá à!                               & You are handsome, but you broke up!             \\ 
                                                         & \multicolumn{2}{l}{FEAR}                             & 5.70                                                                                                      & Tao sợ thật sự.                                    & I'm really scared.                              \\
                                                         & \multicolumn{2}{l}{DISGUST}                          & 4.46                                                                                                      & Diễn viên già vãi -.-                              & The actor is so old -.-                         \\ 
                                                         & \multicolumn{2}{l}{OTHER}                            & 18.66                                                                                                     & Có thể bán lại cho tớ không?                       & Could you sell it for me?                       \\ \hline
\multirow{7}{*}{\textbf{UIT-VSFC}}                       & \multirow{3}{*}{Sentiments}     & POSITIVE            & 49.80                                                                                                     & Slide, giáo trình đầy đủ.                      & Slide and syllabus are full.                    \\ 
                                                         &                                 & NEGATIVE            & 45.80                                                                                                     & Thầy chép bảng nhiều.                              & The lecturer writes on the board a lot.         \\ 
                                                         &                                 & NEUTRAL             & 4.40                                                                                                      & Tăng cường thiết bị..                               & Strengthen equipment.                           \\ \cline{2-6} 
                                                         & \multirow{4}{*}{Topics}         & LECTURER            & 71.70                                                                                                     & Thầy chép bảng nhiều.                              & The lecturer writes on the board a lot.          \\
                                                         &                                 & PROGRAM    & 18.70                                                                                                     & Slide, giáo trình đầy đủ.                      & Slide and syllabus are full.                    \\ 
                                                         &                                 & FACILITY          & 4.20                                                                                                      & Tăng cường thiết bị.                               & Strengthen equipment.                           \\ 
                                                         &                                 & OTHER              & 4.70                                                                                                      &    Hài lòng về tất cả.                        & Satisfied about it all.                         \\ \hline
\multicolumn{1}{l}{\multirow{3}{*}{\textbf{HSD-VLSP}}} & \multicolumn{2}{l}{CLEAN}                            & 91.49                                                                                                     & Hôm nay trời nắng đẹp.                             & It is sunny today.                              \\ 
\multicolumn{1}{l}{}                                   & \multicolumn{2}{l}{OFFENSIVE}                        & 5.02                                                                                                      & vkl.                                              & Cuss.                                           \\ 
\multicolumn{1}{l}{}                                   & \multicolumn{2}{l}{HATE}                             & 3.49                                                                                                      & Nói đến vậy mà cũng không hiểu. Đồ ngu.                & Saying that without understanding. Idiot.       \\ \hline
\end{tabular}

\end{adjustbox}
\caption{Overview statistics of the three Vietnamese social media textual datasets.}
\end{table*}
\end{otherlanguage*}

\section{\textbf{Methodology}}

\begin{figure}[H]
\includegraphics[scale=0.6]{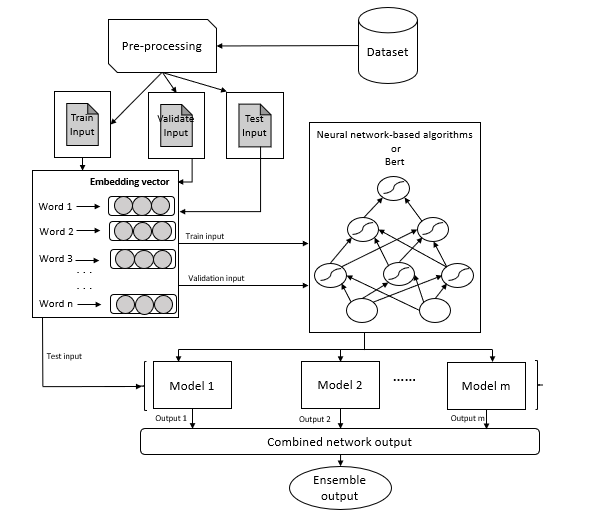}
\caption{Overview of the ensemble model combining neural network-based model for Vietnamese social media datasets}\label{fig:ensemble}
\end{figure}

In this task, we use several pre-processing techniques and deep learning models (CNN  \cite{kim2014convolutional}, LSTM \cite{hochreiter1997long} their variants, and BERT \cite{devlin2018bert}) to apply on the datasets. Each model has its own strength on each label. Therefore, after implementing single models, we propose a simple and efficient ensemble approach combining the best neural network models to improve the classifier performance. Figure \ref{fig:ensemble} shows an overview of the ensemble model for Vietnamese social media text.

\subsection{Pre-processing}
We implement several pre-processing techniques for Vietnamese comments/posts, as follows:
All words are converted to lowercase. URLs and non-alphabetic characters are removed (includes number, excess whitespace).
Also, we use a dictionary to convert abbreviations and slang words to normal to increase the performance of pre-trained embedding as well as use vnTokenizer to tokenize words. (Word segmentation is an important task that helps models achieve better results, especially for the Vietnamese language. More and more research works on this; these works use different methods and achieve high results, such as using CRFs and SVMs \cite{nguyen2006vietnamese}, Hybrid Approach \cite{huyen2008hybrid}. In this study, we use vnTokenizer \cite{huyen2008hybrid} to tokenize words. Table 2 shows some examples in our dictionary.
\begin{otherlanguage*}{vietnamese}
\begin{table}[!htbp]
\centering
\begin{tabular}{lrr}
\hline
\multicolumn{1}{c}{\textbf{No.}} & 
\multicolumn{1}{c}{\textbf{Abbreviation}} & \multicolumn{1}{c}{\textbf{Vietnamese meaning}} \\ 
\hline
1 	& “Chờiiii”	& “Trời”\\ 
2 	 &“vklllll”	& “vkl”\\ 
3 	 &“chetme”	& “chết mẹ”\\ 
4 	 &“kbh”	& “không bao giờ”\\ \hline
\end{tabular}
\caption{Examples of word normalization dictionaries}
\end{table}
\end{otherlanguage*}

\subsection{Word Embeddings}
In this task, we use three pre-trained embeddings for the Vietnamese language, which are currently assessed as the best performance embeddings. Those are fastText \cite{grave2018learning}, Word2vec \cite{word2vecvn_2016} with 300 dim and 400 dim. fastText \cite{grave2018learning} is used for many languages and brought good results on social media data. Therefore, we use fastText \cite{grave2018learning} for the Vietnamese language in this task. Word2vec \cite{word2vecvn_2016} is Word2vec model for the Vietnamese language and is trained on Vietnamese texts. Because they are Word2vec for the Vietnamese language, thus recognizing vast Vietnamese words and brings quite good results on datasets.

\subsection{Convolutional Neural Network (CNN)}
In this task, we use CNN \cite{kim2014convolutional} to classify emotions for Vietnamese text. CNN is an advanced deep learning model, which includes hidden layers such as pooling layers, convolutional layers, fully connected layers, and normalization layers. CNN achieved good results for this document classification task in general and the best on UIT-VSMEC \cite{ho2019emotion} and HSD-VLSP \cite{luu2020comparison}.
\subsection{Long Short-Term Memory (LSTM) and Variants}
Like CNN \cite{kim2014convolutional}, LSTM \cite{hochreiter1997long} is also a modern classification method. This method is strong in classification problems, and most of it has achieved high-performance classification results. Therefore, in this task, we decide to choose it to compare with other classification models. LSTM is a special kind of RNN \cite{medsker1999recurrent}. LSTM’s network architecture includes memory cells and ports that allow the storage or retrieval of information. We also use Bi-LSTM with Bidirectional \cite{650093}, BiLSTM can learn more contextual information extracted from two directions.
\subsection{Gated Recurrent Units (GRU)}
GRU is a variant of RNN and also achieves high results in classification problems. GRU has only two gates, that is the reset gate and the update gate. GRU does not have memory cells like LSTM \cite{hochreiter1997long}. It has only the outputs to make decisions and to inform the next steps. These two gates filter the information from the inputs of the cell and provide an output that satisfies both the criteria of storage past information and the ability to make current decisions most accurately.
\subsection{BERT (Bidirectional Encoder Representations for Transformers)}
BERT \cite{devlin2018bert} is a transfer learning model that achieves state-of-the-art results on natural language processing tasks. As opposed to directional models, which read the text input sequentially (left-to-right or right-to-left), the Transformer encoder reads the entire sequence of words at once. The model can learn the context of a word based on all of its surroundings. Fortunately, BERT is provided with a version for multilingual, including Vietnamese. We apply a BERT-based classifier with a pre-trained multilingual representation into the social media datasets in Vietnamese.
\subsection{Our Proposed Ensemble Method}
Ensemble approaches were very effective in previous studies \cite{huynh2020,tran2020}. We also find that each model has a high classification performance on certain labels so we create an ensemble model based on our implemented neural network models. For each sentence in the test set, the ensemble model selected the label according to the different models' votes. In the case of equal votes, the label was selected based on the model had the best classify performance for that label. Algorithm 1 shows the steps of the ensemble model.
\begin{algorithm}
\caption{- \textbf{An ensemble method for Vietnamese social media textual data}
\label{alg:dsw}}

\begin{flushleft}
\textbf{Input:} A social media text $T$ and m top-performance models $M_{i}$ ($0<i<m+1$).\\
\textbf{Output: } Returning the label $C$ such that our proposed ensemble method predicts.   
\end{flushleft}

\begin{algorithmic}
\Procedure{Ensemble method based on voting}{}
\State{Initialize dictionary D containing the votes of each label.}
\For{i = 1 to m}
    \State $L_{i}$ = $M_{i}(T)$
    \If{$M_{i}(T)$ not in D}
    \State{{$D[L_{i}] = 1$}}
    \Else
    \State{{$D[L_{i}] = D[L_{i}] + 1$}}
    \EndIf
    
    \If{len(D) != m }
    \State{{C = $L_{i}\in D.keys()$, where $L_{i}$ is the label with the highest number of vote.  }}
    \Else
    \State{{C is the label predicted by the model with the best performance.}}
    \EndIf
\EndFor
\State \textbf{return} $C$
\EndProcedure
\end{algorithmic}
\end{algorithm}

\section{\textbf{Experiments}}

\subsection{Evaluation Metric}
To make comparisons with previous studies, we used the corresponding measure based on previous studies' measure, including Weighted F1-score for UIT-VSMEC \cite{ho2019emotion}, Micro F1-score for UIT-VSFC \cite{van2018uit}, with HSD-VLSP dataset, the organization has stopped  providing a test set to ensure fairness in the next competition, so when the contest ended, we could not submit the predicted results to test the accuracy. When doing this study because of the lack of test set, we decide to use the k-fold cross-validation (k = 5) method and Macro F1-score \cite{luu2020comparison} to evaluate the models we applied to HSD-VLSP. 

\subsection{Experimental Settings}
In this study, we experiment with deep learning models that are evaluated as superior in the field of text classification, including LSTM, CNN, BERT, and their variants for the datasets. First, we clean the input data by removing expressions, numbers, special characters, and all words are converted to lowercase. Second, we use some pre-trained embedding that published for the Vietnamese language to represent words into vectors before putting them into deep learning models, including Word2vec and fastText's pre-trained embedding. In this step, with pre-trained embedding was trained on word-separated data, we separate the words for the datasets by the vnTokenizer \cite{huyen2008hybrid}. fastText's pre-trained embedding demonstrates a more efficient representation of words with social datasets. However, we realize that there are some specific words to the classification classes, although represented in pre-trained embedding, they are abbreviated or written in slang, so they can not be recognized and ignored. Therefore, we try to optimize the performance of pre-trained embedding by creating word dictionaries to normalize the above words to the normal form so that they can be represented correctly. Whether it would improve our models' performance? The example of our dictionaries is presented in Table 2.

We also experiment with different values of parameters around the average value of sentences' length. To evaluate whether the pre-processing of data has a significant effect on the classification results, we perform experiments on the processed data and the original data except for the UIT-VSFC dataset because it is cleaned before labeling. However, for the  UIT-VSMEC, the pre-processing do not improve the efficiency compared to the original data. Therefore, we only show the results of the models on the original data for UIT-VSFC and UIT VSMEC dataset and results of the models on the pre-processing data for HSD-VLSP dataset in this study. After experimenting with single models, we combine our models using the max-voting method and model priority as presented in Algorithm 1.

\subsection{Experimental Results}
Our experiments achieve positive results, creating dictionaries that enhance our single model's performance 2\%. With our single models, CNN model has shown its outstanding performance when combining with fastText's pre-trained embedding, which achieves the best results on social media commentary datasets. With the UIT-VSMEC dataset, the CNN model with three layers reach the best efficiency. Conversely,  with the HSD-VLSP dataset, the CNN model has the best performance with five layers due to its larger size.  Besides, BERT has higher performance than the other models when applying to the UIT-VSFC dataset. With a combination of the single models' strengths, our ensemble model has achieved high results. Table 3 shows our results through experiments performed.

\begin{table*}[!htbp]
\centering
\begin{tabular}{lrrrrr}
\hline
\multicolumn{1}{c}{\multirow{2}{*}{\textbf{Model}}} & \multicolumn{2}{c}{\textbf{UIT-VSMEC}}                                               & \multicolumn{2}{c}{\textbf{UIT-VSFC}}                                          & \multicolumn{1}{c}{\multirow{2}{*}{\textbf{HSD-VLSP}}} \\ \cline{2-5}
\multicolumn{1}{c}{}                                & \multicolumn{1}{c}{\textbf{Seven labels}} & \multicolumn{1}{c}{\textbf{Six labels}} & \multicolumn{1}{c}{\textbf{Sentiments}} & \multicolumn{1}{c}{\textbf{Topics}} & \multicolumn{1}{c}{}                                   \\ \hline
BiLSTM + fastText (300 dim)                           & 56.93                                      & 64.18                                    & 91.53                                    & 88.12                                & 85.19                                                   \\ 
BiLSTM + Word2vec (300 dim)                           & 56.16                                      & 62.24                                    & 90.84                                    & 88.53                                & 78.99                                                   \\ 
BiLSTM + Word2vec (300 dim)                           & 56.40                                      & 60.56                                    & 91.18                                    & 88.25                                & 79.18                                                   \\ \hline
LSTM + fastText (300 dim)                             & 52.12                                      & 59.02                                    & 91.06                                    & 87.42                                & 85.56                                                   \\ 
LSTM + Word2vec (300 dim)                             & 50.69                                      & 57.27                                    & 90.90                                    & 86.79                                & 76.26                                                   \\ 
LSTM + Word2vec (400 dim)                             & 49.32                                      & 56.58                                    & 90.58                                    & 87.36                                & 80.84                                                   \\ \hline
GRU + fastText (300 dim)                              & 52.88                                      & 59.45                                    & 90.77                                    & 88.02                                & 84.95                                                   \\ 
GRU + Word2vec (300 dim)                              & 54.02                                      & 59.70                                    & 90.74                                    & 87.71                                & 77.84                                                   \\ 
GRU + Word2vec (300 dim)                              & 54.30                                      & 59.75                                    & 90.87                                    & 87.68                                & 78.38                                                   \\ \hline
CNN + fastText (300 dim)                              & \textbf{59.87}                             & \textbf{66.54}                           & 90.42                                    & 87.68                                & \textbf{85.74}                                          \\ 
CNN + Word2vec (300 dim)                              & 32.65                                      & 33.76                                    & 89.16                                    & 85.94                                & 79.01                                                   \\ 
CNN + Word2vec (400 dim)                              & 23.76                                      & 47.33                                    & 89.79                                    & 86.16                                & 81.63                                                   \\ \hline
BERT (Based-multilingual-cased)                       & 49.22                                      & 55.97                                    & \textbf{92.51}                           & \textbf{89.67}                       & 65.11                                                   \\ \hline
Our proposed model                                    & \textbf{65.79}                             & \textbf{70.99}                           & \textbf{92.79}                           & \textbf{89.70}                       & \textbf{86.96}                                          \\ \hline
\end{tabular}
\caption{F1-score performances of models on the test sets of various Vietnamese social media textual datasets.}
\end{table*}

\subsection{Comparison With Previous Studies}
The results of our single models as well as the ensemble model are better than those of previous studies that were conducted on the same data set. We follow the same metrics and test set that previous studies have used to make similar comparisons. Table 4, 5 and 6 show the best results we achieved compared to previous studies. 
\begin{center}



\begin{table*}[!htbp]
\centering
\begin{tabular}{lrr}
\hline
\multicolumn{1}{c}{\multirow{2}{*}{\textbf{Model}}} & \multicolumn{2}{c}{\textbf{F1-score (\%)}}                                           \\ \cline{2-3} 
\multicolumn{1}{c}{}                                & \multicolumn{1}{c}{\textbf{Seven labels}} & \multicolumn{1}{c}{\textbf{Six labels}} \\ \hline
CNN + Word2vec \cite{ho2019emotion}                                       & 59.74                                      & 66.34                                    \\ \hline
CNN + fastText (Our implementation)                   & \textbf{59.87}                             & \textbf{66.54}                           \\ \hline
Our proposed ensemble (GRU + CNN + BiLSTM + LSTM)     & \textbf{65.79}                             & \textbf{70.99}                           \\ \hline
\end{tabular}
\caption{The comparison with previous studies on UIT-VSMEC.}
\end{table*}

\begin{table*}[!htbp]
\centering

\begin{tabular}{lrr}
\hline
\multicolumn{1}{c}{\multirow{2}{*}{\textbf{Model}}} & \multicolumn{2}{c}{\textbf{F1-score (\%)}}                                     \\ \cline{2-3} 
\multicolumn{1}{c}{}                                & \multicolumn{1}{c}{\textbf{Sentiments}} & \multicolumn{1}{c}{\textbf{Topics}} \\ \hline
BiLSTM + Word2vec \cite{8606837}                                       & 92.03                                    & 89.62                                \\ 
LD + SVM \cite{8573351}                                             & 92.20                                    & -                                    \\ \hline
BERT (Our implementation)                             & \textbf{92.51}                           & \textbf{89.67}                       \\ \hline
Our proposed ensemble (BERT + CNN + BiLSTM + LSTM)    & \textbf{92.79}                           & 89.38                                \\ \hline
Our proposed ensemble (BERT + CNN + BiLSTM)           & 92.13                                    & \textbf{89.70}                       \\ \hline
\end{tabular}
\caption{The comparison with previous studies on UIT-VSFC.}
\end{table*}

\begin{table*}[!htbp]
\centering
\begin{tabular}{lr}
\hline
\multicolumn{1}{c}{\textbf{Model}}             & \multicolumn{1}{c}{\textbf{F1-score (\%)}} \\ \hline
Text-CNN \cite{luu2020comparison}                                         & 83.04                                       \\ 
Logistic regression$^*$ \cite{huu2019automated}                              & 61.97                                       \\ 
Logistic regression + Random Forest + Extra Tree$^*$ \cite{van1991nlp}  & 58.88                                       \\ 
VDCNN, TextCNN, LSTM, LSTMCNN, SARNN$^*$ \cite{nguyen2019vais}              & 58.45                                       \\ 
BiLSTM$^*$ \cite{do2019hate}                                            & 56.28                                       \\  \hline

CNN + fastText (Our implementation)              & \textbf{85.74}                              \\ \hline
Our proposed ensemble (CNN + BiLSTM + LSTM)      & \textbf{86.96}                              \\ \hline
   
\end{tabular}
\caption{The comparison with previous studies on HSD-VLSP. $^*$ indicates that the result is evaluated on a test set of the VLSP shared task 2019. Others use k-fold cross-validation to evaluate the model (k=5) following the study \cite{luu2020comparison}.}
\end{table*}
\end{center}
\section{\textbf{Result Analysis}}

Through the experiment on the UIT-VSMEC dataset, we find that the classification model got a high accuracy for SADNESS, and ENJOYMENT labels. The labels ENJOYMENT and SADNESS are the two labels that account for a high proportion in the dataset, so the correct classification rate for these labels is quite high. However, each model has its strengths in labels. CNN accomplish the highest accuracy classification rate on ENJOYMENT (73.70\%) and is relatively accurate for the rest. BiLSTM has the best effect at DISGUST (65.90\%) and SADNESS (63.79\%) label. Although LSTM and GRU do not have superior results compared to other models on each label, they play an essential role in voting in the ensemble model. The ensemble model improves the predictions in the ENJOYMENT label from 73.70\% to 75.64\%, the SADNESS label from 63.79\% to 67.20\%, and from 63.04\% to 71.73\% for the FEAR label.

In the UIT-VSFC dataset, with the sentiments field, we find that the classifying model achieve a high right classifying rate on the NEGATIVE and POSITIVE label for the sentiment classification task. These two labels occupy approximately the same rate and are much higher than the NEUTRAL label. Therefore, when classifying, the NEUTRAL label is mistaken for the two, the classification results on the NEUTRAL label are low. With sentences that are too long, sentences that express many emotions in the sentence or sentences that are too short, lacking in subjects makes the model difficult to recognize the feelings of the sentence. The highest accuracy of the NEGATIVE and POSITIVE label is 95.95\% and 94.71\% when implementing BERT model. BiLSTM model achieves the highest performance for NEUTRAL label at 41.31\% while CNN and LSTM have the classification results almost equal to the remaining models. Our ensemble model improves the accuracy of the NEGATIVE label from 95.95\% to 97.01\%. 

For the topic classification task, correct classification was high on the LECTURER and FACILITY labels. The LECTURER label has the highest rate in the dataset with 71.70\%, so the exact classification model is easy to explain. The LECTURER label and FACILITY label have the highest accuracy by using the BiLSTM model with 95.72\% and 93.79\%. The highest performance of PROGRAM and OTHER label is 81.29\% and 48.42\% when implementing BERT. The FACILITY label only accounts for the lowest rate in the dataset with 4.30\%. This label contains words related to tools, facilities in the school. Hence, it was easy to identify and difficult to be confused with other labels. Although FACILITY label took up a small percentage, this label reached a high rate of correct categorization. The OTHER label is the one with the lowest accuracy rating because it is difficult to distinguish from the others. Our ensemble model does not increase the classification performance of each label superior to the single model, but it increases the model's overall prediction rate.
\\
In the HSD-VLSP dataset, we find that the CLEAN label achieve the highest classification result, which is also the highest percentage in the dataset (accounting for 97.82\%). The other two labels tend to be mistaken for the CLEAN label due to the effect of CLEAN was too high. The words that appear a lot in the CLEAN label, the classification model would default to their positive meaning, so when using these words in other meaningful sentences (hate, offensive), the model tend to categorize them into a positive sense. Moreover, HATE and OFFENSIVE labels show the relative same emotional state, which is confusing in the classification process, resulting in a low correct classification rate on these two labels. Through each fold, the classification results of the models change. In particular, the CNN model brings the most stable accuracy through each fold, so overall through five folds this model achieves the highest accuracy. CNN accomplish the highest accuracy classification rate on CLEAN at 99.42\% and OFFENSIVE at 68.60\%. The HATE label has the highest accuracy by using the LSTM model with 85.10\%. Through each fold, LSTM shows superiority when classifying on HATE labels, so when ensemble, we give priority to the LSTM model to classify on label HATE. Therefore, the ensemble model helps to increase the accuracy on this label from 85.10 \% to 85.39 \%. Besides, the ensemble model makes the model's overall prediction rate increase.

\section{\textbf{Conclusion and Future Work}}
In this study, we propose a simple but effective ensemble method for social media text on the three different datasets: HSD-VLSP, UIT-VSFC, and UIT-VSMEC. These datasets are multi-label datasets with various labels depending on each dataset. Our ensemble model accomplish outstanding results higher than the deep neural network, as well as previous studies. The combination of many different models' strengths has achieved higher results than single models. In UIT-VSMEC, we achieve 65.79\% of F1-score. With the UIT-VSFC dataset, our results are 92.79\% in sentiment classification task and 89.70\% in the topic classification task in terms of F1-score. Finally, in HSD-VLSP dataset, we reach 86.96\% of the F1-score.

Besides, our single models also bring better results than previous studies. With single models, we find that, with small datasets, pre-processing does not bring expected results, the UIT-VSMEC dataset gain better performances when skipping the pre-processing step. On the contrary, with large datasets, pre-processing data has a good effect on our models, like data from the VLSP shared task. Also, with specific data domains, pre-trained word embeddings on the corresponding data domain will be more effective. Besides, improving word representation by dictionaries also helps to improve the efficiency of the models. Our dictionaries improve pre-trained embedding performance and increase 2\% the effectiveness of our single models. This study indicate that the BERT model has been highly effective for Vietnamese language classification. However, with the use of pre-trained word embeddings training on Wikipedia data, it does not achieve the expected results when applying to the social media data like UIT-VSMEC and UIT-VSFC. With the datasets related to Vietnamese comments on social media, the CNN model has outperformed other single models. Choosing the number of layers is essential, with small datasets such as the UIT-VSMEC dataset, enhancing layers does not bring high efficiency but only makes the model more complicated. In contrast, the addition of layers reach excellent performance for large datasets such as the HSD-VLSP dataset.

For future works, it is necessary to improve the performance of the task, obtaining better results on the datasets, by studying other neural-based models. Besides, test each model with many different parameters to find more suitable parameters. It is implementing experiments of various data processing methods as well as improving coverage of word embeddings on the datasets to gain better results.



\end{document}